\documentclass{article}

\usepackage{arxiv}

\usepackage{appendix}

\usepackage[utf8]{inputenc} 
\usepackage[T1]{fontenc}    
\usepackage{hyperref}       
\hypersetup{
  colorlinks,
  citecolor=magenta,
  linkcolor=magenta,
  urlcolor=magenta
  }
\usepackage{url}            
\usepackage{booktabs}       
\usepackage{amsfonts}       
\usepackage{nicefrac}       
\usepackage{microtype}      
\usepackage{graphicx}
\usepackage[round]{natbib}
\usepackage{doi}

\usepackage{amsthm}
\newtheorem{theorem}{Theorem}
\newtheorem{lemma}[theorem]{Lemma}
\newtheorem{corollary}[theorem]{Corollary}

\newtheorem{definition}{Definition}

\usepackage{color}
 \usepackage{graphicx}
\usepackage{url}
\usepackage{amsmath}
\usepackage{wrapfig}
\usepackage{amsfonts}

\usepackage{mathrsfs}

\usepackage{mathtools, nccmath}

\DeclarePairedDelimiter{\roundop}\lfloor\rceil
\DeclarePairedDelimiter{\quantop}{[\![}{]\!]}
\DeclarePairedDelimiter\floor{\lfloor}{\rfloor}

\usepackage{epsfig}
\usepackage{graphicx}
\usepackage{amsmath}
\usepackage{amssymb}

\usepackage{multirow}

\usepackage{subfigure}
\usepackage{xspace}

\usepackage{tikz}

\usepackage{float}

\usepackage{xcolor,colortbl}

\usepackage{epsfig}
\usepackage{graphicx}
\usepackage{amsmath}
\usepackage{amssymb}

\usepackage{physics}

\title{Reclaiming Residual Knowledge: A Novel Paradigm to Low-Bit Quantization}

\usepackage{authblk}

\renewcommand*{\Affilfont}{\normalsize\normalfont}
\usepackage{orcidlink}

\newsavebox\affbox
\author[$\dag$,$\ddag$]{R\'{o}is\'{i}n Luo (Jiaolin Luo)\,\orcidlink{0000-0002-5365-0379}\,}
\author[*]{Alexandru Drimbarean}
\author[$\dag$,$\ddag$]{James McDermott\,\orcidlink{0000-0002-1402-6995}\,}
\author[$\dag$,$\ddag$]{Colm O'Riordan}
\affil[$\dag$]{%
  \savebox\affbox{\Affilfont SFI Centre for Research Training in Artificial Intelligence, Dublin, D02 FX65, Ireland
  }%
  \parbox[t]{\wd\affbox}{\protect\centering SFI Centre for Research Training in Artificial Intelligence, Dublin, D02 FX65, Ireland}} 
\affil[$\ddag$]{University of Galway, Galway, H91 TK33, Ireland}
\affil[*]{Tobii Corporation, Galway, Ireland}
\date{}

\renewcommand{\headeright}{postprint}
\renewcommand{\undertitle}{A postprint, accepted by The 35\textsuperscript{th} British Machine Vision Conference (BMVC 2024)}
\renewcommand{\shorttitle}{\textbf{CoRa}: Reclaiming Quantization Residual Knowledge}

\hypersetup{
pdftitle={Reclaiming Residual Knowledge: A Novel Paradigm to Low-Bit Quantization},
pdfsubject={Computer Vision and Pattern Recognition (cs.CV); Artificial Intelligence (cs.AI)},
pdfauthor={R\'{o}is\'{i}n Luo},
pdfkeywords={Low-Bit Quantization, Architecture Search, Neural Combinatorial Optimization, Optimization},
}

\newcommand{\ie}{\textit{i}.\textit{e}.}
\newcommand{\eg}{\textit{e}.\textit{g}.}

\begin{document}
\maketitle


\begin{abstract}
This paper explores a novel paradigm in low-bit (\ie~4-bits or lower) quantization, differing from existing state-of-the-art methods, by framing optimal quantization as an architecture search problem within convolutional neural networks (ConvNets). Our framework, dubbed \textbf{CoRa} (Optimal Quantization Residual \textbf{Co}nvolutional Operator Low-\textbf{Ra}nk Adaptation), is motivated by two key aspects. Firstly, quantization residual knowledge, \ie~the lost information between floating-point weights and quantized weights, has long been neglected by the research community. Reclaiming the critical residual knowledge, with an infinitesimal extra parameter cost, can reverse performance degradation without training. Secondly, state-of-the-art quantization frameworks search for optimal quantized weights to address the performance degradation. Yet, the vast search spaces in weight optimization pose a challenge for the efficient optimization in large models. For example, state-of-the-art BRECQ necessitates $2 \times 10^4$ iterations to quantize models. Fundamentally differing from existing methods, \textbf{CoRa} searches for the optimal architectures of low-rank adapters, reclaiming critical quantization residual knowledge, within the search spaces smaller compared to the weight spaces, by many orders of magnitude. The low-rank adapters approximate the quantization residual weights, discarded in previous methods. We evaluate our approach over multiple pre-trained ConvNets on ImageNet. \textbf{CoRa} achieves comparable performance against both state-of-the-art quantization-aware training and post-training quantization baselines, in $4$-bit and $3$-bit quantization, by using less than $250$ iterations on a small calibration set with $1600$ images. Thus, \textbf{CoRa} establishes a new state-of-the-art in terms of the optimization efficiency in low-bit quantization. Implementation can be found on \url{https://github.com/roisincrtai/cora_torch}.
\end{abstract}


\section{Introduction}

\label{sec:intro}

ConvNets \citep{li2021survey,awais2023foundational} are favored as vision foundation models, offering distinct advantages such as the inductive bias in modeling visual patterns \citep{wang2024theoretical,d2019finding,hermann2020origins}, efficient training, and hardware-friendliness  \citep{mauricio2023comparing}. Network quantization is indispensable in enabling efficient inference when deploying large models on devices with limited resources \citep{han2015deep,neill2020overview,mishra2020survey,liang2021pruning}. Representing floating-point tensors as integers significantly reduces the computational requirements and memory footprint.

Yet, low-bit quantization often leads to severe performance degradation \citep{choukroun2019low,gholami2022survey,li2023model}. For example, the standard accuracy of a \textit{resnet18}, pretrained on ImageNet \citep{deng2009imagenet}, plummets to a mere $1.91\%$ from $67.32\%$, with 4-bit weight-only quantization (WOQ), using \textit{min-max} clipping \citep{nagel2021white}. 
To tackle this issue, two research lines are undertaken: quantization-aware training (QAT) and post-training quantization (PTQ) \citep{gholami2022survey,nagel2020up,guo2018survey}.

QAT methods seek the optimal quantized weights during the training process to minimize performance degradation. Despite their promising performance, the substantial computational and data requirements pose major challenges in deployment efficiency. For instance, the state-of-the-art PACT \citep{choi2018pact} entails a minimum of $10^{8.12}$ iterations with $1.2M$ training samples to converge on ImageNet. Additionally, empirical evidence shows that QAT methods often yield very limited performance at low-bit quantization due to optimization difficulty \citep{nagel2022overcoming,liu2019learning,esser2019learned}. PTQ methods, \eg~AdaRound \citep{nagel2020up} and BRECQ \citep{li2021brecq}, overcome these limitations by reconstructing the optimal quantized weights of pre-trained models, with optimization on small calibration sets; these potentially reduce computational and data requirements.

Notably, both state-of-the-art QAT and PTQ methods quantize models by optimizing within weight spaces. Their optimization efficiencies are substantially hindered by the vast dimensions of search spaces. For example, a \textit{resnet50} contains over 2.5M trainable parameters \citep{he2016deep}, suggesting a search space of the dimension of $\mathbb{R}^{25,000,000}$. The state-of-the-art PTQ method BRECQ needs at least $2 \times 10^4$ iterations to converge for a \textit{resnet50} pre-trained on ImageNet.

\begin{figure*}[t]

  \centering
 \includegraphics[width=0.8\textwidth,]{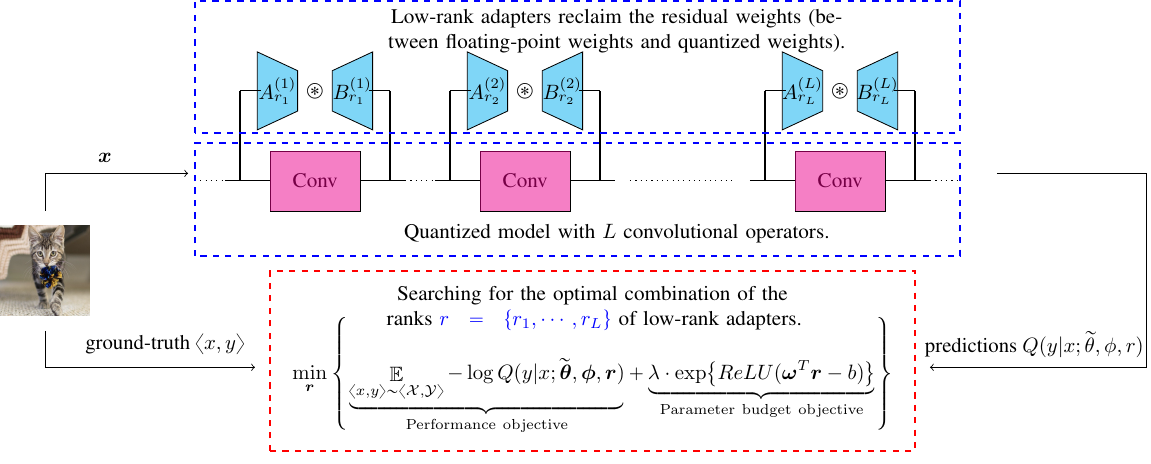}

  \caption{\textbf{CoRa} framework: Searching for the optimal adapters, reclaiming the quantization residual knowledge, instead for the optimal quantized weights. The low-rank convolutional adapter at the $l$-th layer $B_{r_l}^{(l)} \circledast A_{r_l}^{(l)}$ is determined by a discrete integer $r_l$.}

\label{fig:overview}
\end{figure*}

This research delves into a question: ``\textbf{Beyond the quantization methods with weight space optimization, does an alternative paradigm exist}?''. Intuitively, the quantization residual knowledge -- namely, the \textit{quantization residual weights} between floating-point weights and quantized weights -- retains vital information lost during the quantization process. This quantization residual knowledge, which has long been overlooked by the research community, holds the potential value that reverses performance degradation without training. Motivated by this perspective, our approach, \textbf{CoRa}, as shown in Figure~\ref{fig:overview}, explores a novel paradigm, differing from state-of-the-art QAT and PTQ methods: \textbf{by seeking the optimal low-rank adapters \citep{hu2021lora}, reclaiming the residual knowledge; thus reversing the performance degradation, and establishing a new state-of-the-art in terms of the optimization efficiency}. 

A low-rank adapter consists of two cascaded convolutional filters (\eg~$A$ and $B$) with significantly lower sizes, which are directly converted from high-rank quantization residual weights. As shown in Figure~\ref{fig:overview}, the $l$-th layer adapter $B_{r_l}^{(l)} \circledast A_{r_l}^{(l)}$, with a low rank $r_l$, is attached to the $l$-th layer convolutional filter, and approximates the quantization residual weights. \textbf{CoRa} seeks the optimal ranks $r=\{r_1, \cdots, r_L\}$ for all adapters. Surprisingly, earlier works \citep{zhang2015accelerating,yang2020learning,hu2021lora,denton2014exploiting,rigamonti2013learning,jaderberg2014speeding,zhong2024convolution} do not address the problem of converting the existing weights of convolutional operators into the weights of the adapters without training. To tackle this problem, we prove a result, as stated in Residual Convolutional Representation Theorem~\ref{theo:res_conv_theorem}.

The search space of the low-rank adapters in a model is significantly smaller by many orders of magnitude compared to the space of weights. For instance, a \textit{resnet50} has $53$ convolutional filters. In this case, the structure of the low-rank adapters is only controlled by $53$ parameters (\ie~$53$ ranks). This suggests that the search space is of dimension $\mathbb{R}^{53}$, smaller by $6$ orders of magnitude than the weight space. Thanks to the smaller search space, \textbf{CoRa} converges within less than $250$ iterations for pre-trained models on ImageNet, yet achieves comparable performance against both state-of-the-art QAT and PTQ baselines.

This research is in the scope of low-bit WOQ and ConvNets. Our contributions are summarized as: 
\begin{itemize}
   
    \item[1] \textbf{CoRa method}. We present an efficient, low-bit, and PTQ framework for ConvNets, by framing optimal quantization as an architecture search problem, to re-capture quantization residual knowledge with low-rank adapters;

    \item[2] \textbf{Neural combinatorial optimization}. We introduce a differentiable neural combinatorial optimization approach, searching for the optimal low-rank adapters, by using a smooth high-order normalized Butterworth kernel;

    \item[3] \textbf{Training-free residual operator conversion}. We show a result, converting the weights of existing high-rank quantization residual convolutional operators to low-rank adapters without training, as stated in Theorem~\ref{theo:res_conv_theorem}.

\end{itemize}

\section{Preliminaries}

\textbf{Dataset and classifier}. Let $\langle \mathcal{X}, \mathcal{Y}\rangle$ be an image dataset where $\mathcal{X}$ denotes images and $\mathcal{Y}$ denotes labels. We use $Q(y|x; \vb*\theta)$ to represent a classifier, where $\theta$ denotes parameters. $Q$ predicts the probability of a discrete class $y$ given image $x$. 

\textbf{Quantization}. We use $\quantop{W}_n$ to denote the $n$-bit quantization of tensor $W$. The \textit{clipping range} refers to the value range in quantization \citep{gholami2022survey}. We use two clipping schemes: (1) \textit{min-max clipping} chooses the minimum and maximum values. (2) \textit{normal clipping} chooses $[\mu-k\cdot \sigma, \mu+k\cdot \sigma]$, where $\mu$ denotes the mean of the tensor, $\sigma$ denotes the standard deviation of the tensor and $k$ determines the range. Details are in Appendix~\ref{app:uniform_quantization}.

\textbf{Kolda mode-$n$ matricization and tensorization}. Let $Z \in \mathbb{R}^{I_1 \times \cdots \times I_N}$ be a $N$-order tensor \citep{kolda2009tensor}. The Kolda mode-$n$ \textit{matricization} of $Z$ \citep{kolda2009tensor}, denoted as $Z_{(n)}$, refers to \textit{unfolding} $Z$ so that: the $n$-th dimension becomes the first dimension, and the remaining dimensions are squeezed as the second dimension by $\Pi_{i \neq n} I_i$. Let $Y \in \mathbb{R}^{I_n \times J}$ ($J=\Pi_{i \neq n} I_i$) be a matrix. The mode-$n$ \textit{tensorization} of $Y$, denoted as $Y_{[n, I_1 \times I_{n-1} \times I_{n+1} \times \cdots \times I_N]}$, refers to \textit{folding} $Y$ into the shape $I_1 \times  \cdots \times I_{N}$. Readers can further refer to the literature \citep{kolda2009tensor,li2018rtensor,zhou2021decoupled}. Details are also provided in Appendix~\ref{app:tensor_product}.

\textbf{Residual convolutional operator}. Let $W \in \mathbb{R}^{m \times n \times k_1 \times k_2}$ be the weights of a convolution operator, where $m$ denotes output channels, $n$ denotes input channels and $k_1 \times k_2$ denotes filter kernel size. We refer to $W$ as \textit{convolutional operator} for brevity. We use $W \circledast \vb*x$ to denote the convolution operation. Convolutional operators are linear operators. We refer to $\Delta \quantop{W}_n := W - \quantop{W}_n$ as \textit{quantization residual operator}, or \textit{residual operator} if without ambiguity.

\begin{theorem}[Residual Convolutional Representation]
\label{theo:res_conv_theorem}
Suppose a singular value decomposition given by:
$(\Delta \quantop{W}_n)_{(1)} = US_rV^T$ ($r=\rank(S_r) $). Then the factorization holds true:
 \begin{align}
    W \circledast \vb*x = \quantop{W}_n \circledast \vb*x + \underbrace{B \circledast A}_{\mathrm{residual~operator}} \circledast \vb*x
\end{align} 
where $A=(S_r^{\frac{1}{2}}V^T)_{[1, r \times 1 \times 1]}$ and $B=(US_r^{\frac{1}{2}})_{[1, n \times k_1 \times k_2]}$. The $B \circledast A$ is referred as $r$-rank \textit{residual operator}. The proof is provided in Appendix~\ref{app:residual_proofs}.
\end{theorem}

\section{Method}
\label{sec:method}

We frame the optimal quantization as an architecture search problem. Suppose a $L$-layer floating-point ConvNet $Q$:
\begin{align}
Q(y|x; \vb*\theta) := Q(y|x; W^{(1)}, \cdots, W^{(L)})    
\end{align}
in which $W^{(l)}$ denotes the parameters of the $l$-th layer and $\vb*\theta := \{ W^{(1)}, \cdots, W^{(L)} \}$. The quantized $Q$ with bit-width $n$ is:
\begin{align}
   Q(y|x; \widetilde{\vb*\theta}) :=
Q(y|x; \quantop{W^{(1)}}_n, \cdots, \quantop{W^{(L)}}_n) 
\end{align}
where $\widetilde{\vb*\theta} := \{ \quantop{W^{(1)}}_n, \cdots, \quantop{W^{(L)}}_n \}$. 

\textbf{Approximating residual knowledge}. According to Theorem~\ref{theo:res_conv_theorem}, in the $l$-th layer, the residual operator $\Delta \quantop{W^{(l)}}_n$ is approximated by a $r_l$-rank residual operator: 
\begin{align}
\label{equ:delta_w_a_b}
W^{(l)} \circledast x - \quantop{W^{(l)}}_n \circledast x = \Delta \quantop{W^{(l)}}_n \circledast x \approx B^{(l)}_{r_l} \circledast A^{(l)}_{r_l} \circledast x .
\end{align}
Notably, the $A^{(l)}_{r_l}$ and $B^{(l)}_{r_l}$ are directly converted from $\Delta \quantop{W^{(l)}}_n$ without training, which is guaranteed by Theorem~\ref{theo:res_conv_theorem}. By approximating the residual operators via Equation~\eqref{equ:delta_w_a_b}, the quantized model is written as:
\begin{align}
Q(y|x; \widetilde{\vb*\theta},\vb*\phi, \vb*r) :=
Q(y|x; \quantop{W^{(1)}}_n + B^{(1)}_{r_1} \circledast A^{(1)}_{r_1}, \cdots, \quantop{W^{(L)}}_n + B^{(L)}_{r_L} \circledast A^{(L)}_{r_L}) 
\end{align}
where $\vb*\phi = \{ B^{(1)}_{r_1} \circledast A^{(1)}_{r_1}, \cdots, B^{(L)}_{r_L} \circledast A^{(L)}_{r_L}\}$ are the low-rank residual operators, and $\vb*r = \{r_1, \cdots, r_L\}$ ($0 \leq r_l \leq R_l$, $r_l \in \mathbb{N}$) are the parameters controlling the ranks of these operators. The implementation is as shown in Figure~\ref{fig:overview}.

\textbf{Discrete combinatorial optimization}. Suppose $r=\{ r_1, \cdots, r_L\}$ are a set with $L$ discrete ranks, controlling the structure of the low-rank adapters in Figure~\ref{fig:overview}. Suppose $R_l$ is the $l$-th layer maximum rank of $r_l$. Formally, the optimization objective is to seek a set of optimal discrete $r$, by maximizing the performance on a calibration set $\langle \mathcal{X}, \mathcal{Y}\rangle$, subject to an adaptation parameter budget constraint: 
\begin{align}
\label{equ:combinatorial_optimization}
\vb*r^* = \arg \min_{\textcolor{blue}{\vb*r}} \quad &\left\{ \mathop\mathbb{E}_{\langle x, y\rangle \sim \langle \mathcal{X}, \mathcal{Y}\rangle} 
-\log Q(y|x; \widetilde{\vb*\theta}, \vb*\phi, \textcolor{blue}{\vb*r})  \right\}  \nonumber\\
     \text{subject to} \quad &\vb*\omega^T\textcolor{blue}{\vb*r} \leq b 
     ,0 \leq \textcolor{blue}{r_i} \leq R_i
      ,~\textcolor{blue}{r_i} \in \mathbb{N}, ~0 \leq b \leq 1
\end{align}
where $\vb*r^*$ denotes the optimal ranks, $b$ denotes normalized maximum adaptation parameter budget (\ie~target budget), and $\vb*\omega := \{\omega_1, \cdots, \omega_L \}$ denotes the \textit{rank normalization coefficients} used to compute the normalized parameter size. The $\omega_l$ is given by: $\omega_l := \frac{1}{R_l} \cdot \frac{\Theta_l}{\sum_{i=1}^L \Theta_i}$ where $\Theta_i$ is the $i$-th layer parameter size, as proved in Appendix~\ref{app:weighting_factors}. The optimization search space size is less than dimension of $\mathbb{R}^{L}$. The only learnable parameters are $r$.

\textbf{Adapter parameter budget constraint}. We limit the amount of the parameters of the adapters by:
\begin{align}    
    \label{equ:budget_constraint_implementation}
    \min\limits_{\textcolor{blue}{\vb*r}} \left\{
    \lambda \cdot \exp{ReLU(\vb*\omega^T \textcolor{blue}{\vb*r} - b)}
    \right\}
\end{align}
where $\lambda \in \mathbb{R}$ is a penalty coefficient. The motivation of using $\exp(\cdot)$ is to obtain non-linear gradients favoring gradient-based optimization, by assigning smaller gradients to smaller ranks, instead of the constant gradients $\omega$. The $ReLU(\cdot)$ is used to stop gradients if the running budget $\omega^T r$ is already below the target budget $b$.

\begin{figure*}[!t]
 

  \centering
  
 %

    \subfigure[singular values $S^{(l)}$]{
	   \begin{minipage}[b]{0.31\textwidth}
	   \includegraphics[width=1\textwidth,]{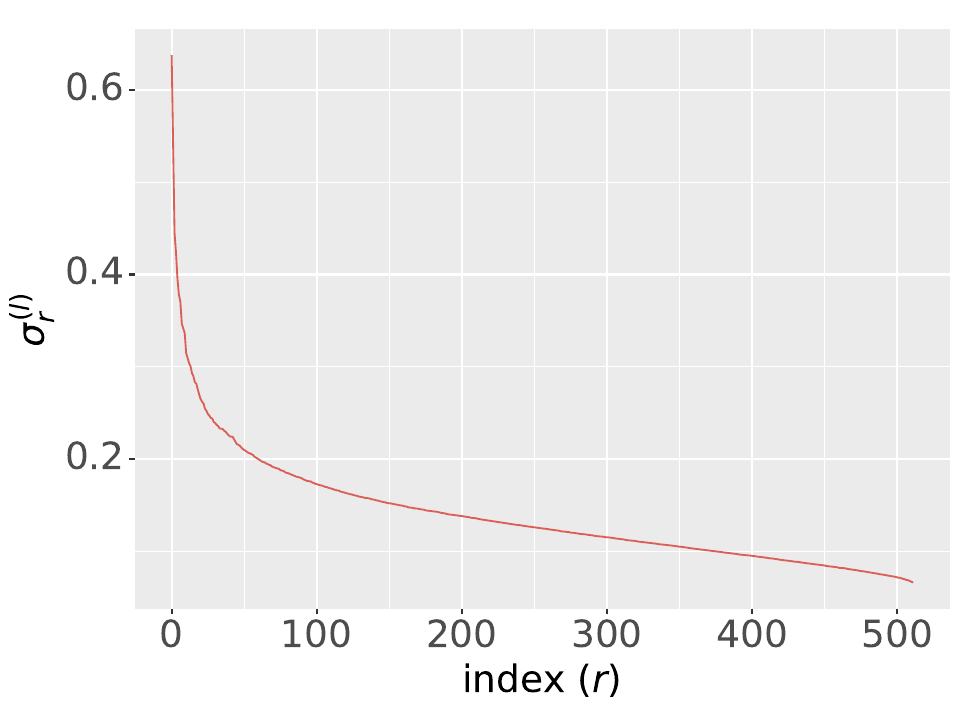}
		\end{minipage}
		\label{subfig:sigmas}
	} \hspace*{\fill}%
    \subfigure[$\Phi(r_l)$ w/ various orders]{
	   \begin{minipage}[b]{0.31\textwidth}
	   \includegraphics[width=1\textwidth,]{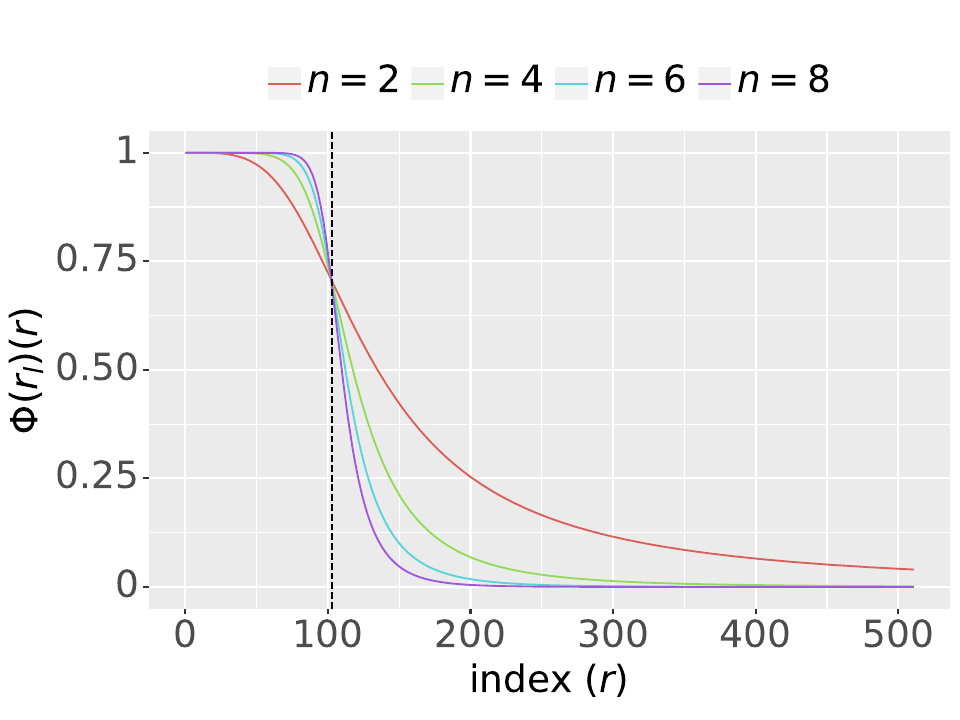}
	   \end{minipage}
		\label{subfig:nbk}
	}\hspace*{\fill}%
    \subfigure[thresholding by $\Phi(r_l) \odot S^{(l)}$]{
	   \begin{minipage}[b]{0.31\textwidth}
	   \includegraphics[width=1\textwidth,]{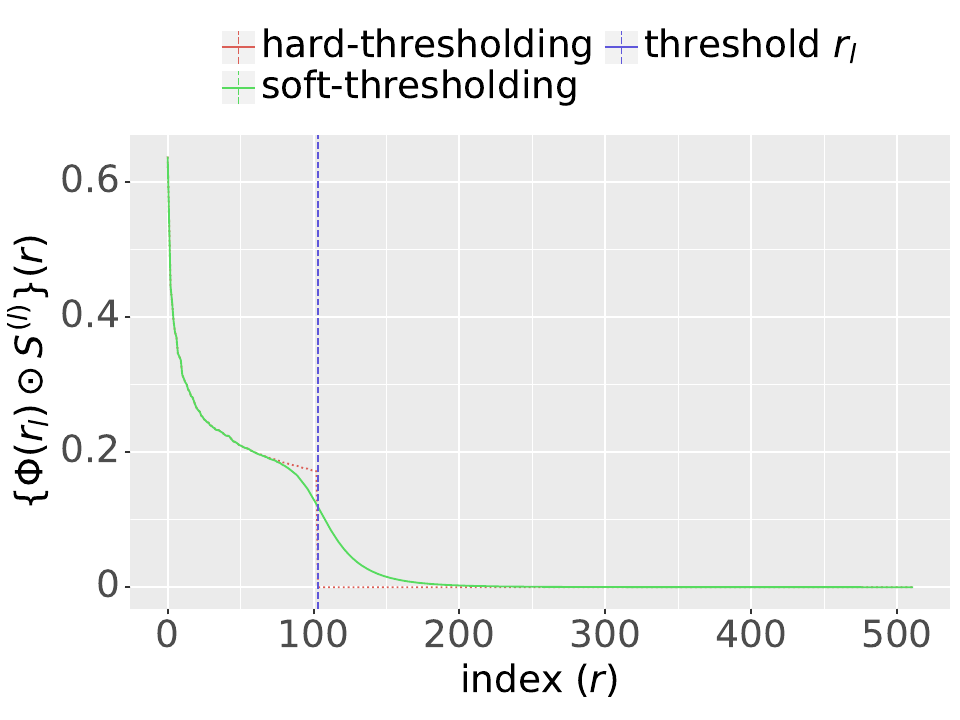}
	    \end{minipage}
		\label{subfig:thresholding}
	}

  
  \caption{Differentiable thresholding with a high-order normalized Butterworth kernel $\Phi(r_l)$. The $S^{(l)}$ is in the $14$-th layer of a pre-trained \textit{resnet18} on \textit{ImageNet}. The cut-off rank $r_l$ is $103$.}
  
 
\label{fig:thresholding_s_matrix}
\end{figure*}

\subsection{Differentiable relaxation}
\label{sec:relaxation}

Equation~\eqref{equ:combinatorial_optimization} is not differentiable with respect to $r$. Solving the discrete combinatorial optimization problem in Equation~\eqref{equ:combinatorial_optimization} often entails iterative algorithms, \eg~evolutionary algorithms (EA) and integer programming (IP) \citep{bartz2014evolutionary,mazyavkina2021reinforcement,wolsey2020integer}. Nevertheless, the huge discrete search spaces remain a significant hurdle. For instance, the number of possible combinations of low-rank adapter sizes in a \textit{resnet50} is above $10^{18}$. 

To enable efficient optimization, firstly, we relax the $r$ in Equation~\eqref{equ:combinatorial_optimization} from discrete integers to continuous values. Secondly, we differentiably parameterize the operations of choosing $r$, by using a high-order normalized Butterworth kernel. With these endeavors, Equation~\eqref{equ:combinatorial_optimization} is differentiable with respect to $r$. We are able to use standard gradient descent algorithms to efficiently optimize (\eg~SGD and Adam).

\begin{figure*}[t]
 

  \centering
  

  

  \subfigure[running budget ($\omega^Tr$)]{
	   \begin{minipage}[b]{0.43\textwidth}
	   \includegraphics[width=1\columnwidth,]{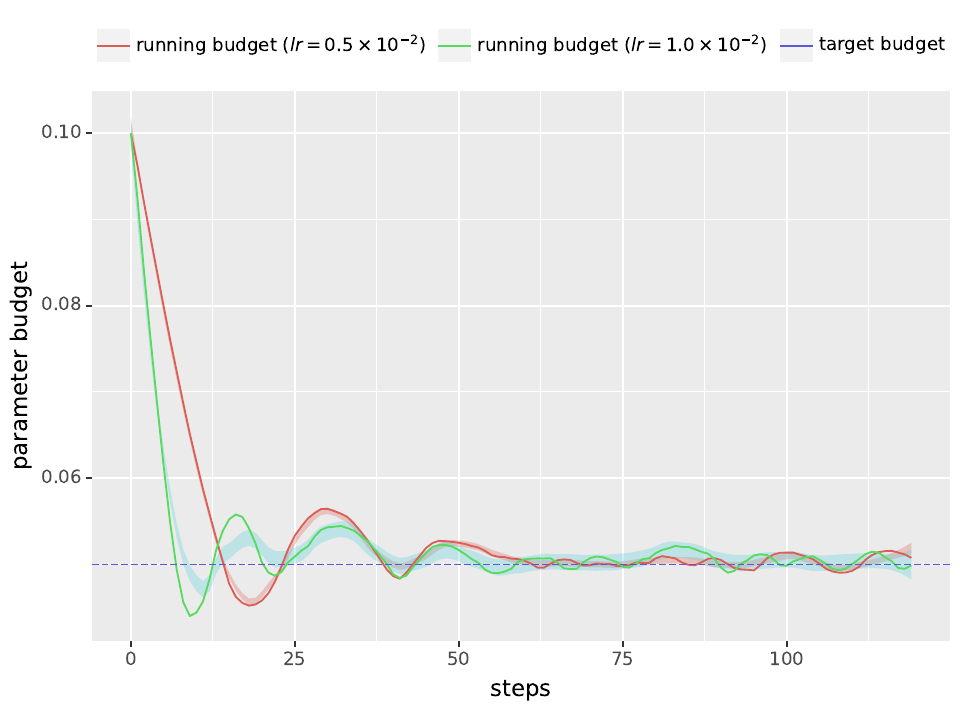}
		\end{minipage}
		\label{subfig:running_budget}
	}\hspace*{\fill}%
 \subfigure[solution]{
	   \begin{minipage}[b]{0.43\textwidth}
	   \includegraphics[width=1\columnwidth,]{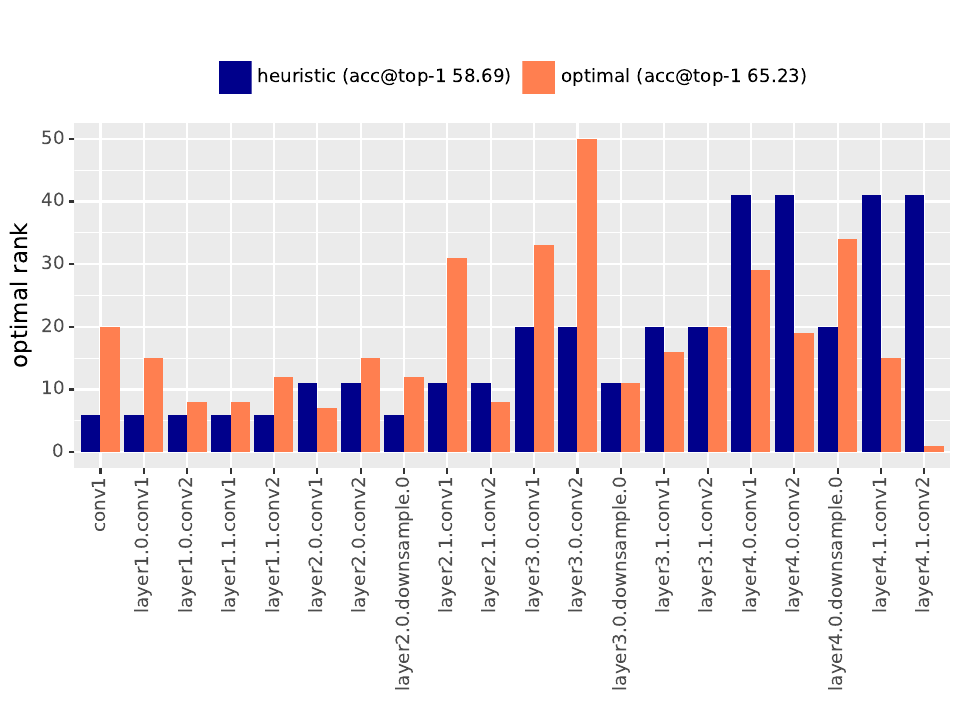}
		\end{minipage}
		\label{subfig:solution}
	}

 
  \caption{Optimization iterations and solution. The experiments are with \textit{resnet18} pre-trained on ImageNet.}


\label{fig:optimization_example}
\end{figure*} 

\subsection{Parameterized differentiable thresholding}
\label{sec:diff_thresholding}

\textbf{Hard thresholding}. Suppose $S^{(l)}$ is the singular value matrix of $\quantop{\Delta W^{(l)}}_n$. Suppose $S_{r_l}^{(l)}$ chooses the $r_l$ largest values of $S^{(l)}$:
\begin{align}
S^{(l)}=diag(\sigma_1^{(1)},\cdots, \sigma_{R_l}^{(l)})  
\quad
S_{r_l}^{(l)}=diag(\sigma_1^{(1)},\cdots, \sigma_{r_l}^{(l)},0, \cdots, 0)
\quad
0\leq r_l \leq R_l.
\end{align}
Formally, choosing the $r_l$ can be formulated as the Hadamard product (\ie~element-wise product) of a thresholding mask matrix $\Phi^*(r_l)$ and $S^{(l)}$ in that: 
\begin{align}
\label{equ:hard_thresholding}
    S_{r_l}^{(l)} = \Phi^*(r_l) \odot S^{(l)}
\qquad 
\Phi^*(r_l)=diag(\underbrace{1,\cdots,1}_{r_l~\mathrm{ones}},\underbrace{0,\cdots,0}_{R_l - r_l~\text{zeros}}).
\end{align}
We refer to Equation \eqref{equ:hard_thresholding} as \textit{hard-thresholding}, which is not differentiable with respect to $r$.

\textbf{Soft thresholding}. We differentiably approximate the \textit{hard-thresholding} with a high-order normalized Butterworth kernel (NBK) \citep{butterworth1930theory}. An $k$-order NBK with a cut-off rank $r_l$ is a vector map $\Phi(r_l): r_l \mapsto [0,1]^{R_l}$ defined by:
\begin{align}
    \label{equ:butterworth_kernel}
    \Phi(r_l) := \left( 
    \frac{1}{\sqrt{1 + (\frac{r}{r_l})^{2k} } } 
    \right)_{r = 1}^{R_l}
    \qquad r_l \ll R_l
\end{align}
where $n$ is the order. Figure~\ref{subfig:sigmas} shows an example of $S^{(l)}$. Figure~\ref{subfig:nbk} shows an example of NBK. Figure~\ref{subfig:thresholding} shows the results of the differentiable thresholding with NBK. 

\textbf{Converting residual operator to low-rank operator}. In the $l$-th layer, we differentiably convert a high-rank residual operator $\Delta \quantop{W^{(l)}}_n$ into a low-rank operator $B_{r_l}^{(l)} \circledast A_{r_l}^{(l)}$ with rank $r_l$, by using Equation~\eqref{equ:butterworth_kernel}:
\begin{align}
    \Delta \quantop{W^{(l)}}_n &\approx B_{r_l}^{(l)} \circledast A_{r_l}^{(l)} \\
    &\approx
     ({U^{(l)}} [ \Phi(r_l) \odot {S^{(l)}}^{\frac{1}{2}} ] )_{[1, n \times k_1 \times k_2]}
     \circledast
    ([ \Phi(r_l) \odot {S^{(l)}}^{\frac{1}{2}}] {V^{(l)}}^T)_{[1, r_l \times 1 \times 1]}.
\end{align}
which is differentiable with respect to $r_l$.

\begin{figure}[!t]
    \centering

    
    \begin{minipage}[t]{0.48\textwidth}
        \centering
        \includegraphics[width=1\textwidth,]{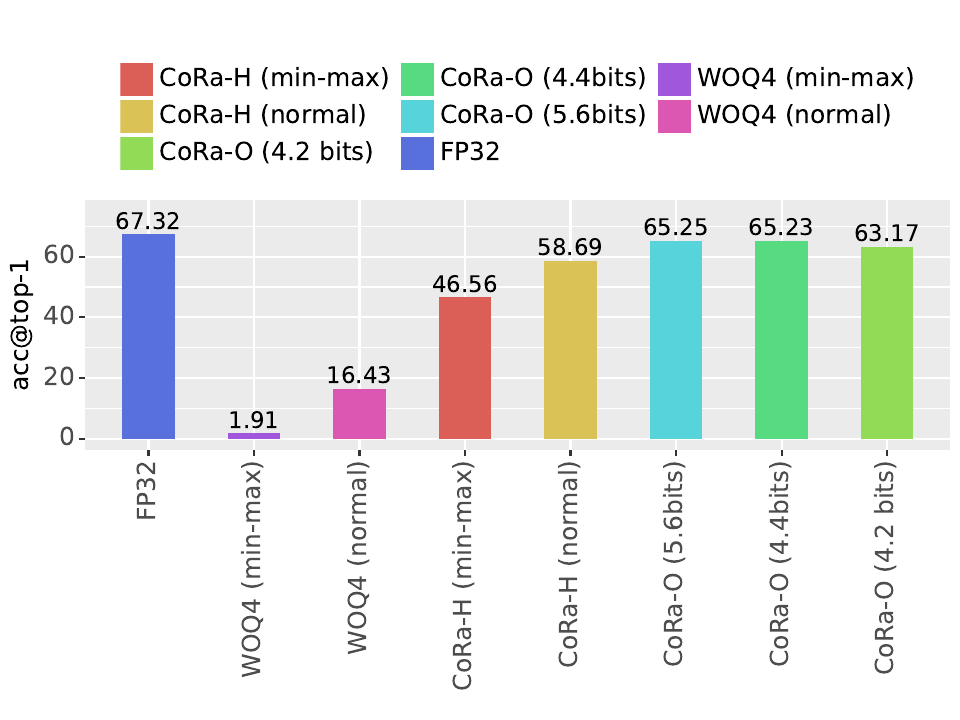}
        \vspace{-2.5em}
        
        \caption{Ablation study on \textit{ImageNet} with \textit{resnet18}. \textbf{CoRa-H}: Heuristic ranks. \textbf{CoRa-O}: Optimal ranks.}
        
        \label{fig:ablation}
    \end{minipage}
    \hfill
    \begin{minipage}[t]{0.48\textwidth}
        \centering
        \includegraphics[width=1\textwidth,]{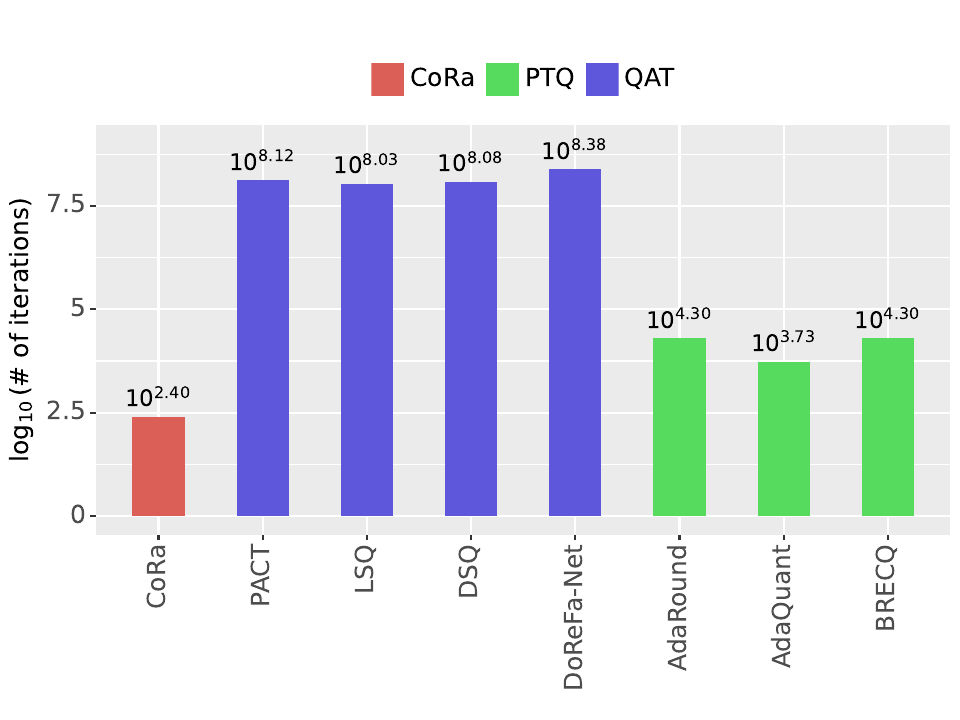}
        \vspace{-2.5em}
        \caption{Optimization efficiency on \textit{ImageNet}. Results are in a logarithmic scale.}
        \label{fig:efficiency}
    \end{minipage}



\end{figure}


\subsection{Neural combinatorial optimization}
\label{sec:neural_comb_optim}

By combining Equation~\eqref{equ:combinatorial_optimization} and Equation~\eqref{equ:budget_constraint_implementation}, the optimization loss is:
\begin{align}
\mathcal{L}(\textcolor{blue}{\vb*r}) :=
\mathop\mathbb{E}_{\langle x, y\rangle \sim \langle \mathcal{X}, \mathcal{Y}\rangle} 
\left\{
-\log Q(y|x; \widetilde{\vb*\theta}, \vb*\phi, \textcolor{blue}{\vb*r}) 
+
\lambda \cdot \exp{ReLU(\vb*\omega^T\textcolor{blue}{\vb*r} - b)}
\right\}.
\end{align}
The optimal $r=\{r_1, \cdots, r_L\}$ are found using gradient descent optimizers, \eg SGD and Adam.

\textbf{Heuristic choice of ranks}. This method serves as a baseline. The $r_l$ is heuristically chosen as $r_l= \lfloor b \cdot R_l \rfloor $, proportionally assigning the $l$-th layer rank according to the budget $b$. For example, suppose the maximum rank at the $l$-th layer is $512$ and budget $b$ is set to $5\%$, the heuristic $r_l$ is chosen as $\lfloor 512 \times 0.05 \rfloor = 25$.


\textbf{Intriguing observation}. Figure~\ref{subfig:running_budget} shows the running budget $\omega^T r$ during the optimization. Figure~\ref{subfig:solution} shows an example of the solution on a \textit{resnet18} on ImageNet. Our analysis in terms of the solutions from a variety of ConvNets suggests that: \textbf{The heuristic choices often overstate the importance of middle to last layers; conversely, optimal solutions underscore the importance of beginning to middle layers}.

\subsection{Tricks for stable optimization}
Stable optimization for the proposed neural combinatorial optimization in Section~\ref{sec:neural_comb_optim} is challenging. We adopt several tricks to numerically stabilize the optimization process.
\begin{itemize}

    \item \textbf{Gradient clipping}. To stabilize the optimization, the solver clips the gradients into the range of $[-0.2, 0.2]$.

    \item  \textbf{Adaptive gradient}. Equation~\eqref{equ:budget_constraint_implementation} gives non-linear gradients with respect to ranks. Smaller ranks have smaller gradients towards zero, while larger ranks have larger gradients. We believe this design favors the stabilization of optimization.

    \item  \textbf{Solution clamping}. The solver clamps the ranges of the solutions after every gradient update, guaranteeing that the rank is not less than $1$ and not greater than the limit $R_l$.

    \item \textbf{Anomaly reassignment}. The solver detects numerical anomalies. If a \textit{NaN} rank value is detected, it is replaced with rank $1$.
\end{itemize}

\section{Experiments}

We conduct experiments from three aspects: (1) ablation study (Section~\ref{sec:exp_ablation}), (2) comparing with state-of-the-art QAT and PTQ baselines (Section~\ref{sec:exp_comparison}), and (3) extensive evaluation (Section~\ref{sec:exp_extensive}). 

\textbf{Reproducibility}. We sample $1600$ images from the \textit{ImageNet} validation set as our calibration set while using the remainder as our validation set. We use \textit{normal clipping} with $k=4.0$ to quantize the main network and \textit{min-max clipping} to quantize adapters. The order $n$ of NBK is set to $4.0$. The penalty coefficient $\lambda$ is set to $1$. The batch size is $32$. The target budget $b$ is set to $5\%$, which results in a $1.25\%$ increase in memory footprint with $8$-bit quantization for low-rank adapters. The optimizer is Adam without weight decay. The learning rate is set to $0.01$. We use a maximum $250$ iterations for all experiments. 

\textbf{Testbed}. All experimental results, including the measured results of floating-point reference accuracy, are conducted on the M2 chip of a MacBook Air, equipped with a GPU of size 24 GiB. Due to the choice of the validation set, in tandem with the random seed and the hardware acceleration implementation in the testbed, the results of reference accuracy are slightly lower compared to the results from pytorch. However, this does not affect the results, we obtain using pytorch, for a fair comparisons with baselines. We report the results that we measured on our own testbed rather than using the results from the literature.

\textbf{Equivalent quantization bit-width}. Let $n$ and $m$ be the quantization bit-widths of the main network and adapters. The equivalent quantization bit-width is given as: $n + m \cdot b$. For example, suppose $n=4$, $m=8$ and $b=5\%$, the equivalent quantization bit-width is $4.4$-bits. The proof is provided in Appendix~\ref{app:equiv_quant_bits}.

\begin{table}[t]

\caption{Top-1 accuracy comparison of low-bit quantization. We use marker $\times$ to indicate that the results are not available. }

\vspace{1em}


\label{tab:imagenet_results}

\centering


\resizebox{0.9\columnwidth}{!}{
\begin{tabular}{c||c||c||c||c|c|c|c||c|c|c}
\hline

\multirow{2}{*}{Model} &\multirow{2}{*}{bits} &\multirow{2}{*}{FP32} & \multirow{2}{*}{CoRa} & \multicolumn{4}{c||}{QAT Baselines} & \multicolumn{3}{c}{PTQ Baselines} \\
\cline{5-11} 

 & & &   & PACT & LSQ & DSQ & DoReFa-Net & AdaRound & AdaQuant & BRECQ \\
\hline
\hline

resnet18 & \multirow{2}{*}{4} & 67.32 & \cellcolor{cyan!20}\textbf{65.23} & 66.08 & 67.89 & 66.99 & 65.99 &65.08 & 65.18 & 66.96 \\

resnet50 && 74.52 & \cellcolor{cyan!20}\textbf{72.85} & $\times$ & 73.84 & 73.39 & $\times$ & 72.81 & 72.80 & 73.88 \\

\hline
\hline

resnet18 & \multirow{2}{*}{3} & & \cellcolor{cyan!20}\textbf{64.50} & 65.31 & 66.13 & $\times$ & $\times$ &64.47 & 55.04 & 66.12 \\

resnet50 && & \cellcolor{cyan!20}\textbf{72.81} & $\times$ & $\times$ & $\times$ & $\times$ & 71.06 & 65.43 & 70.87 \\

\hline
\hline

\rowcolor{cyan!40}
\# of iterations && $10^{8.38}$ &$\leq 10^{2.40}$ & $10^{8.12}$ & $10^{8.03}$ & $10^{8.08}$ & $ 10^{8.38}$ & $10^{4.30}$ & $10^{3.73}$ & $10^{4.30}$ \\

\hline
training size && $1.2M$ & $1600$ & $1.2M$ & $1.2M$ & $1.2M$ & $1.2M$ & 2048 & 1000 & 1024 \\

\hline

\end{tabular}
} 

\vspace{-1em}
\end{table}

\subsection{Ablation study}
\label{sec:exp_ablation}

We conduct ablation experiments to show the design considerations in: (1) \textit{normal} clipping is better than \textit{min-max} clipping, (2) the results with optimal ranks outperform the results with heuristic choices, and (3) quantizing residual adapters with $8$-bits does not affect performance. The results are shown in Figure~\ref{fig:ablation}. 

Intriguingly, we can quantize adapters while the performance remains almost unchanged. This can significantly reduce the amount of extra parameters which are used to retain residual knowledge.

\subsection{Comparing with baselines} 
\label{sec:exp_comparison}

We compare our method against both state-of-the-art QAT and PTQ baselines. We choose four QAT baselines: PACT \citep{choi2018pact}, LSQ \citep{esser2019learned}, DSQ \citep{gong2019differentiable}, and DoReFa-Net \citep{zhou2016dorefa}. We choose three PTQ baselines: AdaRound \citep{nagel2020up}, AdaQuant \citep{hubara2020improving}, and BRECQ \citep{li2021brecq}. We quantize the \textit{resnet18} and \textit{resnet50} (pre-trained on \textit{ImageNet}) with $4$-bits and $3$-bits quantization.

\textbf{Top-1 accuracy}. Our results achieve comparable performance against the baselines. The results are shown in Table~\ref{tab:imagenet_results}. \textbf{Optimization efficiency}. Our method is more efficient by many orders of magnitude than state-of-the-art baselines. The results are shown in Figure~\ref{fig:efficiency} and Table~\ref{tab:imagenet_results}. Notably, our method uses only $250$ iterations with very minimum extra parameter cost. We have established a new state-of-the-art in terms of optimization efficiency.

\subsection{Extensive evaluation} 

\label{sec:exp_extensive}

\begin{figure*}[t]

  \centering

  
  \includegraphics[trim=0cm 1.2cm 0cm 0cm, width=1\textwidth,]{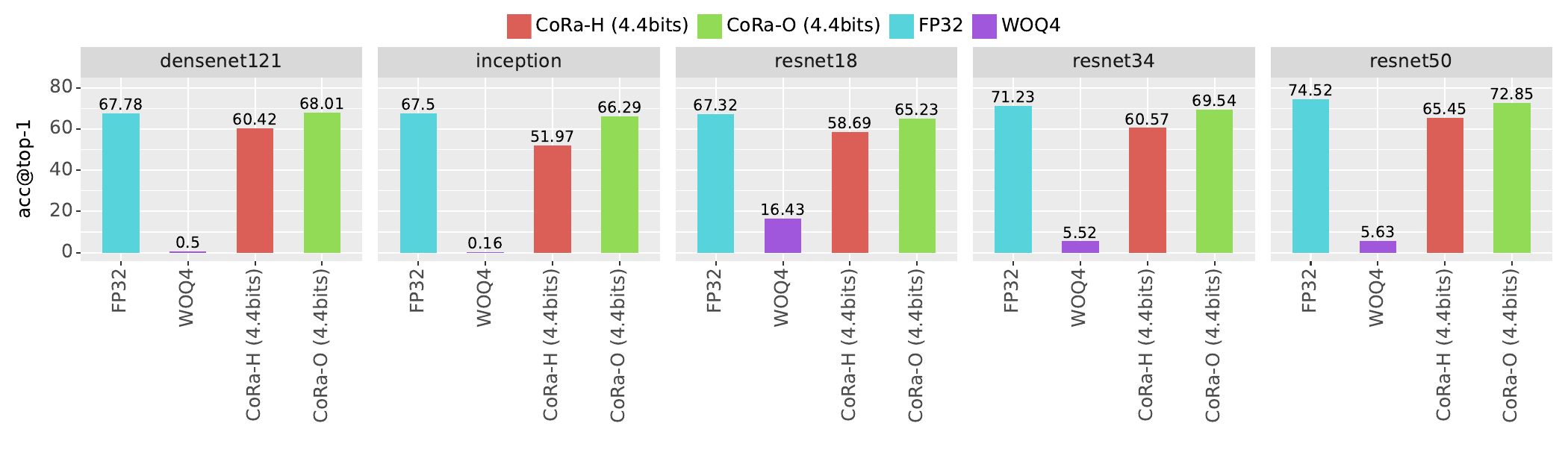}

 
  \caption{Top-1 accuracy of multiple vision architectures on \textit{ImageNet} with $4$-bit quantization. 
  }
  

\label{fig:results}
\end{figure*}

We show the results of extensive performance evaluation over multiple image classifiers pre-trained on ImageNet in Figure~\ref{fig:results}. Our results achieve comparable performance against the floating-point reference models with the differences within $2.5\%$. The full solutions are provided in Appendix~\ref{app:full_solutions}.






\section{Related work}
\label{sec:related_work}

\textbf{Low-rank convolutional operator approximation}. Low-rank approximation of convolutional operators is promising in accelerating the computations \citep{li2021survey}. However, convolution operations are not matrix multiplications. Conventional low-rank approximation, \eg~LoRa \citep{hu2021lora} and QLoRa \cite{dettmers2024qlora}, fails to approximate convolutional operators. Relatively few works in the literature have explored this problem. \citeauthor{denton2014exploiting} decompose filters into the outer product of three rank-$1$ filters by optimization \citep{denton2014exploiting}. \citeauthor{rigamonti2013learning} use rank-$1$ filters to approximate convolutional filters by learning \citep{rigamonti2013learning}. \citeauthor{jaderberg2014speeding} reconstruct low-rank filters with optimization, by exploiting the significant redundancy across multiple channels and filters \citep{jaderberg2014speeding}. A recent work, Conv-LoRA, approximates filters with the composed convolutions of two filters for low-rank fine-tuning on ConvNets \citep{zhong2024convolution}. However, Conv-LoRA does not solve the problem of converting existing operators to low-rank operators without training. Previous works need to reconstruct low-rank filters by learning, thus they do not satisfy our needs. \textbf{CoRa} uses Theorem~\ref{theo:res_conv_theorem} to convert existing residual operators into low-rank operators without training.

\section{Future work}

\textbf{CoRa} introduces a novel paradigm in low-bit quantization and demonstrates significant optimization efficiency, with a new state-of-the-art result, as shown by our experiments compared to baselines. This paper exclusively investigates this paradigm on ConvNets. Future research will aim to explore this paradigm further from three aspects: (1) enhancing the performance of existing quantization methods (\eg~QAT and PQT) by reclaiming the residual knowledge using \textbf{CoRa}; (2) extending this paradigm to architectures beyond ConvNets, such as transformers; and (3) broadening the scope to more diverse tasks, including large vision models (LVMs) and large language models (LLMs).

\section{Conclusions}
We explore a novel paradigm, in optimal low-bit quantization, differing from existing state-of-the-art methods, by re-framing the problem as an architecture search problem, of optimally reclaiming quantization residual knowledge. Thanks to significantly smaller search spaces of adapters, our method is more efficient yet achieves comparable performance against state-of-the-art baselines. \textbf{CoRa} has established a new state-of-the-art in terms of the optimization efficiency.

\section*{Acknowledgements}
This publication has emanated the grant support from research conducted with the financial support of Science Foundation Ireland under grant number 18/CRT/6223. Financial support has also been provided by XPeri Corporation (Galway) and Tobii Corporation (Galway). For the purpose of Open Access, the author has applied a CC BY public copyright licence to any Author Accepted Manuscript version arising from this submission. We thank researchers Fatemeh Amerehi and Laura O'Mahony, from the University of Limerick (Ireland), for their proofreading. Specially, we extend our gratitude to the reviewers for their constructive comments, which have significantly enhanced the quality of our research.

\bibliographystyle{unsrtnat}
\bibliography{ref}  

\newpage
\section*{Appendix}
\appendix

\section{Uniform quantization}
\label{app:uniform_quantization}

We formally introduce uniform quantization, which refers to the integer representations of floating-point tensors by taking the quantization intervals uniformly \citep{gholami2022survey,guo2018survey}.

Suppose: 
\begin{align}
\roundop{\cdot}: \mathbb{R}^{I_1 \times I_2 \times \cdots \times I_N} \mapsto \mathbb{Z}^{I_1 \times I_2 \times \cdots \times I_N}
\end{align}
is an element-wise rounding operator in tensor space such as $round(\cdot)$, $floor(\cdot)$ or $ceil(\cdot)$ in pytorch \citep{ketkar2021introduction}. 

\textbf{Quantization operator}.
Suppose:
\begin{align}
\quantop{\cdot}_{n}: \mathbb{R}^{I_1 \times I_2 \times \cdots \times I_N} \mapsto \mathbb{Z}^{I_1 \times I_2 \times \cdots \times I_N}    
\end{align}
is a $n$-bit quantization operator which sends tensors from floating-point representations to $n$-bit integer representations. 

\textbf{De-quantization operator}.
We define the de-quantization operator as:
\begin{align}
 \quantop{\cdot}_{n}^{-1}: \mathbb{Z}^{I_1 \times I_2 \times \cdots \times I_N} \mapsto \mathbb{R}^{I_1 \times I_2 \times \cdots \times I_N}.   
\end{align}

Let $\vb*x \in \mathbb{R}^{I_1 \times I_2 \times \cdots \times I_N}$ be some floating-point tensor. Let $[\alpha, \beta] \subset \mathbb{R}$ be the quantization representation range (\ie~quantization clipping range) where $\alpha,\beta \in \mathbb{R}$. Choosing the clipping range constant $[\alpha, \beta]$ is often referred as `calibration' \citep{gholami2022survey}. Let $s \in \mathbb{R}$ be some scale constant determined by quantization bit-width and representation range $[\alpha, \beta]$: \begin{align}
    s(n;\alpha;\beta) = \frac{\alpha - \beta}{2^n - 1}.
\end{align}

Suppose $z \in \mathbb{Z}$ denotes the integer representation zero point (\ie~quantization bias), the quantization operator $\quantop{\cdot}_{n}$ can be formulated as: 
\begin{align}
    \vb*x^* = \quantop{\vb*x}_{n} \stackrel{def}{=} \roundop{ \frac{\vb*x}{s(n;\alpha;\beta)}  } - z
\end{align}
which represents $\vb*x$ into the range:
\begin{align}
    [ \roundop{\frac{\alpha}{s(n;\alpha,\beta)}} -z, \roundop{\frac{\beta}{s(n;\alpha,\beta)}} - z] \subset \mathbb{Z}.
\end{align}

The choices of $z$ determine two schemes for uniform quantization: (1) Symmetric quantization scheme and (2) asymmetric quantization scheme (\ie~affine quantization) \citep{gholami2022survey,jacob2018quantization,krishnamoorthi2018quantizing}. 

For example, the scheme $z:=\frac{\alpha}{s(n;\alpha,\beta)}$ is dubbed as `asymmetric quantization scheme' or `affine quantization scheme' where the floating-point representation zero-point is mapped to the bias $z$. The scheme $z:=0$ is dubbed as `symmetric quantization scheme' where the floating-point representation zero-point is mapped to zero.

Accordingly, the de-quantization operator $\quantop{\cdot}_{n}^{-1}$ can be formulated as:
\begin{align}
    \vb*x = \quantop{\vb*x^*}_{n}^{-1} \stackrel{def}{=} s(n;\alpha,\beta) \cdot (\vb*x^* + z).
\end{align}

\newpage
\section{Mode-$n$ tensor product}
\label{app:tensor_product}

Let $Z \in \mathbb{R}^{I_1 \times I_2 \times \cdots \times I_N}$ be a $N$-order tensor where $I_i$ denotes the $i$-th dimension size. A `fiber' refers to the vector created by fixing $N-1$ dimensions. A `slice' refers to the matrix created by fixing $N-2$ dimensions. 

The mode-$n$ matricization of tensor $Z$ is also known as tensor `unfolding'. For example, a tensor with shape $8 \times 16 \times 3 \times 3$ can be unfolded into a matrix with shape $8 \times 144$. The mode-$n$ matricization of tensor $Z$ is denoted as $Z_{(n)}$ by arranging the mode-$n$ fibres of $Z$ as columns. The mode-$n$ product is known as tensor-matrix product. The mode-$n$ product of tensor $Z$ and matrix $Y \in \mathbb{R}^{J \times I_n}$ is defined as $Z \times_n Y \stackrel{def}{=} Y Z_{(n)}$. 

Accordingly, we also define the inverse mode-$n$ tensorization (\ie~`folding') as the inverse operation and denote it as $Z_{[n, J_1 \times J_2 \times \cdots \times J_K]}$ where $J_1 \times J_2 \times \cdots \times J_K$ denotes the folding dimensions of the unfolded dimensions in $Z_{(n)}$. Readers can refer to the literature \citep{kolda2009tensor} for details.

\newpage
\section{Proof: Residual convolutional representation}
\label{app:residual_proofs}

We aim to prove:
 \begin{align}
    W \circledast \vb*x = \quantop{W}_n \circledast \vb*x + \underbrace{B \circledast A}_{\mathrm{residual~operator}} \circledast \vb*x
\end{align} 
as stated in Theorem~\ref{theo:res_conv_theorem}.

However, the convolutional operations are not matrix multiplications. We can not directly use the results such as singular value decomposition (SVD). Strictly proving the Theorem~\ref{theo:res_conv_theorem} demands some efforts. We use the knowledge from tensor algebra to show that Theorem~\ref{theo:res_conv_theorem} holds.

\begin{definition}[\textbf{Unfolding operator}]
\label{def:unfolding}
Let:
\begin{align}
\mathbb{T}(k_1 \times k_2,s_1 \times s_2): \mathbb{R}^{n \times w \times h} \mapsto \mathbb{R}^{n\cdot k_1 \cdot k_2 \times w' \times h'}
\end{align}
be the `unfolding' operation which arranges some input with shape `$n \times w \times h$' to shape `$n \cdot k_1 \cdot k_2 \times w' \times h$' for convolution operation with respect to kernel size $k_1 \times k_2$ and stride size $s_1 \times s_2$. In particular, if the stride size is $1 \times 1$, we simplify the notation as:
\begin{align}
\mathbb{T}(k_1 \times k_2).    
\end{align}
Suppose $\vb*x \in \mathbb{R}^{n \times w \times h}$. For example, the operator with stride $s_1 \times s_2$ and no padding is defined by:
\begin{align}
    \mathbb{T}(k_1 \times k_2, s_1 \times s_2)(\vb*x)(c, u,v) := \vb*x(c\mod{(n \cdot k_1 \cdot k_2)}, 
     \floor{\frac{u}{s_1} }, 
    \floor{\frac{v}{s_2} })
\end{align}
where $c, u$ and $v$ are indices. Clearly:
\begin{align}
 \mathbb{T}(1 \times 1)(\vb*x) \equiv \vb*x   
\end{align}
holds true as a particular case. Readers can refer to the implementation of the operation \textit{unfold} in pytorch.
\end{definition}

\begin{lemma}[\textbf{Tensor mode-$n$ product factorization}]
    \label{lemma:mode_n_product_decomp}
    Let $Z \in \mathbb{R}^{I_1 \times I_2 \times \cdots \times I_N}$ be some tensor. Let $A \in \mathbb{R}^{R \times I_n}$ and $B \in \mathbb{R}^{J \times R}$ be matrices. Below identity holds:
    \begin{align}
        Z \times_{n} A \times_{n} B \equiv Z \times_{n} (BA).
    \end{align}
    Readers can refer to literature \citep{kolda2009tensor} for the proof.
\end{lemma}

\begin{lemma}[\textbf{Convolution mode-$n$ representation}]
\label{lemma:conv_mode_n_product} 
Let $W \in \mathbb{R}^{m \times n \times k_1 \times k_1}$ be the weights of some `$k_1 \times k_1$' 2D convolutional operator where $m$ denotes output channels and $n$ denotes input channels. Let $\vb*x \in \mathbb{R}^{n \times w \times h}$ be some input with size $w \times h$ and channels $n$. According to the definition of 2D convolution, the convolution $W \circledast \vb*x$ can be represented as:
\begin{align}
    W \circledast \vb*x = \underbrace{\mathbb{T}(k_1 \times k_2)(\vb*x)}_{n \cdot k_1 \cdot k_2 \times w' \times h'} \times_1 W_{(1)}.
\end{align}

\end{lemma}

\begin{theorem}[\textbf{Convolution factorization}]
\label{theo:conv_separation}
Let $Z \in \mathbb{R}^{m \times n \times k_1 \times k_2}$ be the weights of some 2D convolutional operator. Suppose:
\begin{align}
 Z_{(1)} = B_{(1)}A_{(1)} \in \mathbb{R}^{m \times n\cdot k_1 \cdot k_2}   
\end{align}
where $A \in \mathbb{R}^{d \times n \times k_1 \times k_2}$ and $B \in \mathbb{R}^{m \times d \times 1 \times 1}$ are the weights of two convolutional operators with kernel sizes `$k_1 \times k_1$' and `$1 \times 1$' respectively. Let $\vb*x \in \mathbb{R}^{n \times w \times h}$ be some input. Using Lemma~\ref{lemma:conv_mode_n_product} and Lemma~\ref{lemma:mode_n_product_decomp}:
\begin{align}
    Z \circledast \vb*x &= \mathbb{T}(k_1 \times k_2)(\vb*x) \times_1 Z_{(1)} \\
    &= \mathbb{T}(k_1 \times k_2)(\vb*x) \times_1 (B_{(1)} A_{(1)}) \\
    &= \mathbb{T}(k_1 \times k_2)(\vb*x) \times_1 A_{(1)}  \times_1 B_{(1)} \\
    &=(A \circledast \vb*x) \times_1 B_{(1)} \\
    &=\mathbb{T}(1 \times 1)(A \circledast \vb*x) B_{(1)} \\
    &= B \circledast A \circledast \vb*x.
    \end{align}
\end{theorem}

\begin{corollary}[\textbf{Convolutional singular value decomposition}]
\label{cor:conv_svd}
Suppose:
\begin{align}
 Z_{(1)} = USV^T = US^{\frac{1}{2}} (S^{\frac{1}{2}}V)^T   
\end{align}
where:
\begin{align}
S^{\frac{1}{2}} \odot S^{\frac{1}{2}} = S.    
\end{align}
Set:
\begin{align}
B_{(1)} := US^{\frac{1}{2}}    
\end{align}
and: 
\begin{align}
A_{(1)} := (S^{\frac{1}{2}}V)^T.
\end{align}
Using Theorem~\ref{theo:conv_separation}:
\begin{align}
    Z \circledast \vb*x = (US^{\frac{1}{2}})_{[1,d \times 1 \times 1]} \circledast (S^{\frac{1}{2}}V)^T_{[1, n \times k_1 \times k_2]} \circledast \vb*x.
\end{align}
\end{corollary}
    
\begin{proof}
We now show that the Theorem~\ref{theo:res_conv_theorem} strictly holds by using Theorem~\ref{theo:conv_separation} and Corollary~\ref{cor:conv_svd}. Suppose $W \in \mathbb{R}^{m \times n \times k_1 \times k_2}$ be the weights of some convolutional operator. Set:
\begin{align}
Z := \Delta \quantop{W}_n \in \mathbb{R}^{m \times n \times k_1 \times k_2}.    
\end{align}
Convolutional operators are linear operators. The convolution quantization residual representation is:
\begin{align}
    W \circledast \vb*x &= \quantop{W}_n \circledast \vb*x + \Delta \quantop{W}_n \circledast \vb*x  \\
    &= \quantop{W}_{n} \circledast \vb*x+ (US^{\frac{1}{2}})_{[1, d \times 1\times 1]} \circledast (S^{\frac{1}{2}}V)^T_{[1, n\times k_1 \times k_2]} \circledast \vb*x.
\end{align}
There is nothing to do. Theorem~\ref{theo:res_conv_theorem} holds as demonstrated.
\end{proof}

\newpage
\section{Rank normalization coefficients}
\label{app:weighting_factors}

Suppose a model with $L$ convolutional filters. Suppose the $l$-th layer has parameter size $\Theta_l$. Suppose the adaptation of the $i$-th layer has parameters $\Xi_i$. The maximum parameter size of the $l$-th adapter is $\Theta_l$. 

The overall size of the adapters is given by:
\begin{align}
    \frac{r_l}{R_l} \cdot \Theta_l.    
\end{align}

Normalizing with respect to the overall model size:
\begin{align}
   \sum\limits_{i=1}^L \Theta_i.
\end{align}

Thus, the running budget is:
\begin{align}
   \frac{r_l}{R_l} \cdot \Theta_l \cdot \frac{1}{\sum\limits_{i=1}^L \Theta_i} \leq b. 
\end{align}

Set:
\begin{align}
    \omega_l = \frac{1}{R_l} \cdot \frac{\Theta_l}{\sum\limits_{i=1}^L \Theta_i}
\end{align}
which is referred as $l$-th layer \textit{rank normalization coefficient}.

\newpage
\section{Equivalent quantization bit-width}
\label{app:equiv_quant_bits}

Suppose the $l$-th layer parameter size $\Theta_l$. The the $l$-th layer adapter size is:
\begin{align}
    \frac{r_l}{R_l} \cdot \Theta_l.
\end{align}

The equivalent quantization bit-width $\xi$ is:
\begin{align}
\label{equ:model_size}
    \frac{\xi}{32} = 
    \frac
    {\frac{n}{32} \cdot \sum\limits_{i=1}^L \Theta_i + \frac{m}{32} \cdot \sum\limits_{i=1}^L \frac{r_l}{R_l} \cdot \Theta_i }
    {\sum\limits_{i=1}^L \Theta_i}.
\end{align}

Simplifying the Equation~\eqref{equ:model_size}:
\begin{align}
    \xi = n + m \cdot b.
\end{align}

\newpage
\section{Full solutions}
\label{app:full_solutions}

We provide full solutions for all experimental models. 

\begin{figure}[H]
 
  \centering

  \includegraphics[width=0.8\columnwidth,]{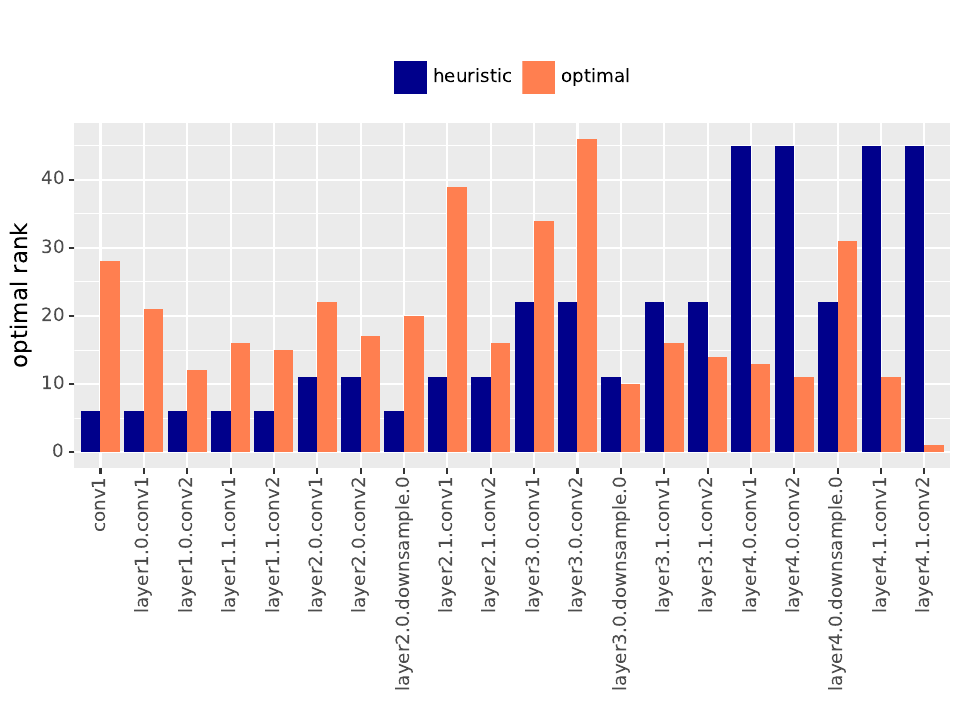}

  \caption{Solution for \textit{resnet18}.}

\end{figure}

\newpage

\begin{figure}[H]

  \centering

  \includegraphics[width=0.8\columnwidth,]{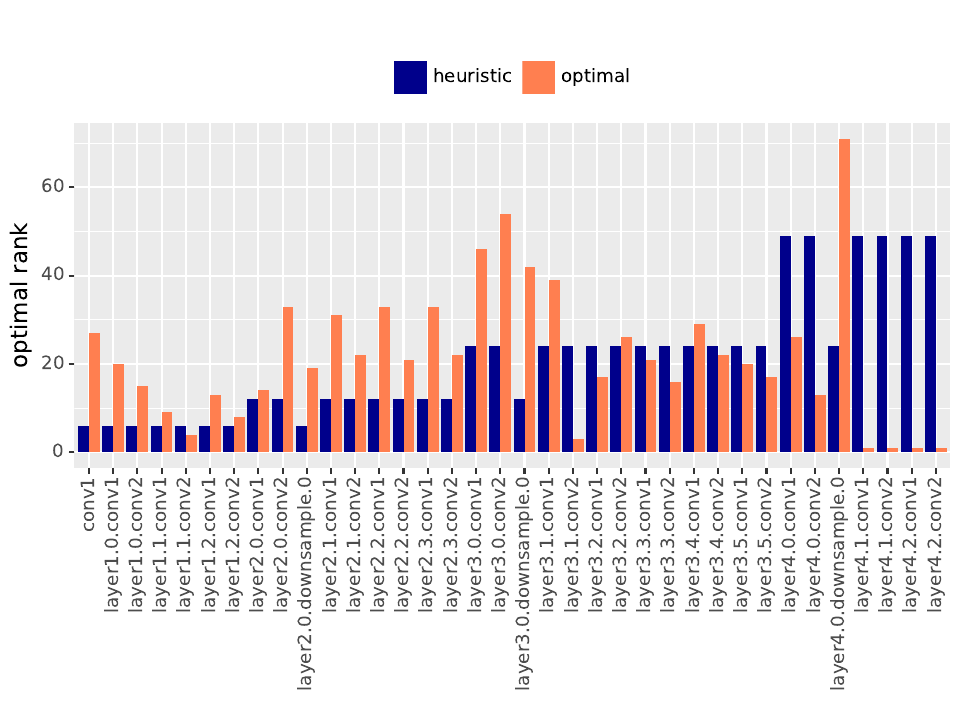}

  \caption{Solution for \textit{resnet34}.}
  

\end{figure}

\newpage

\begin{figure}[H]

  \centering
 
  \includegraphics[width=0.8\columnwidth,]{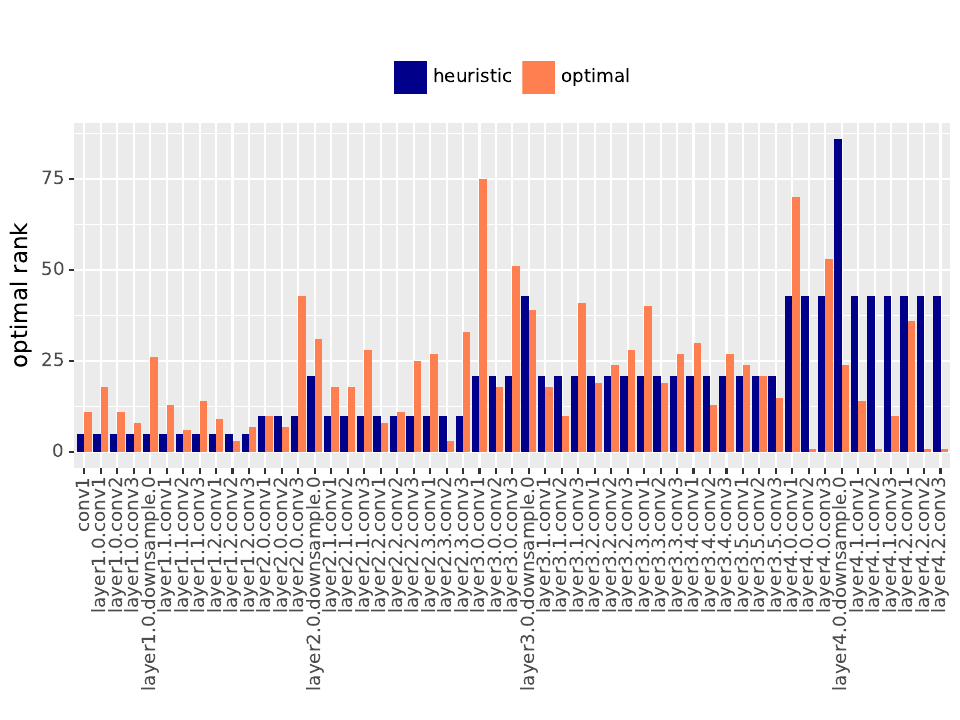}

  \caption{Solution for \textit{resnet50}.}
  

\end{figure}

\newpage

\begin{figure}[H]

  \centering
 
  \includegraphics[width=0.8\columnwidth,]{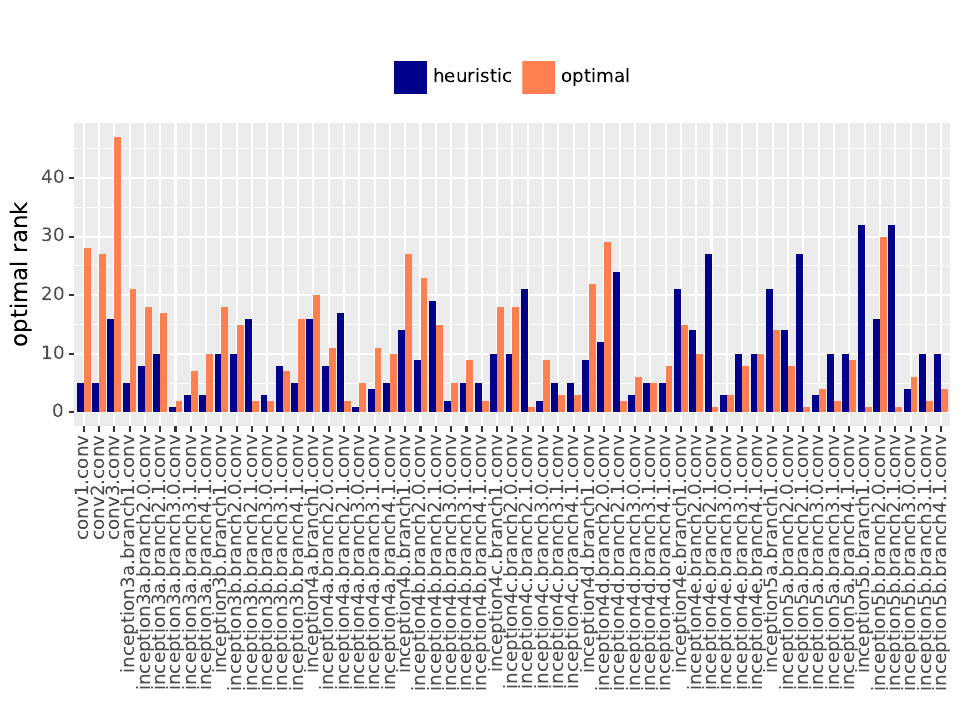}

  \caption{Solution for \textit{inception}.}
  

\end{figure}

\newpage

\begin{figure}[H]

  \centering

  \includegraphics[width=0.8\columnwidth,]{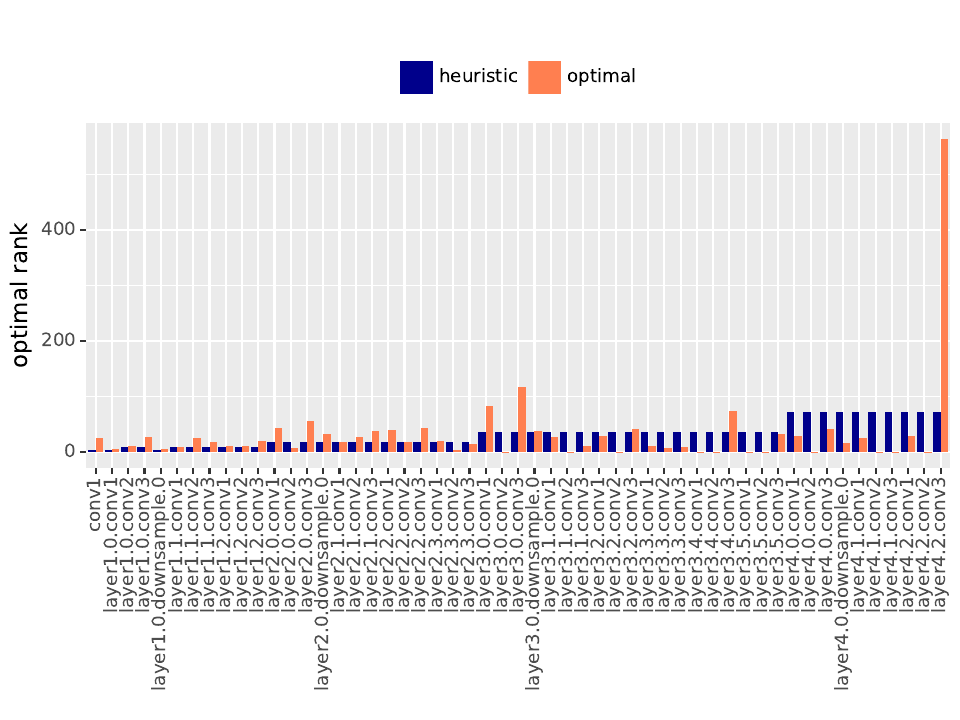}

  \caption{Solution for \textit{wide\_resnet50\_2}.}
  

\end{figure}

\end{document}